\documentclass{article}
\usepackage[switch]{lineno} % <-- switch alternates line numbers between columns
\usepackage{caption} % in preamble

\usepackage{adjustbox} % in the preamble

\usepackage{microtype}
\usepackage{graphicx}
\usepackage{subfigure}
\usepackage{booktabs} % for professional tables
\usepackage{enumitem}
\usepackage{tabularx,booktabs}

\usepackage{hyperref}
\usepackage{array,tabularx,booktabs,makecell}

\usepackage[accepted]{icml2025}

\usepackage{amsmath}
\usepackage{amssymb}
\usepackage{mathtools}
\usepackage{amsthm}
\usepackage{multirow}

\usepackage[capitalize,noabbrev]{cleveref}

\theoremstyle{plain}

\theoremstyle{definition}

\theoremstyle{remark}

\usepackage[textsize=tiny]{todonotes}

\begin{document}

\twocolumn[
\icmltitle{PLUTO-4: Frontier Pathology Foundation Models}




\begin{icmlauthorlist}

% Example ICML-style author block
\icmlauthor{Harshith Padigela}{lead}
\icmlauthor{Shima Nofallah}{core}
\icmlauthor{Atchuth Naveen Chilaparasetti}{core}
\icmlauthor{Ryun Han}{core}
\icmlauthor{Andrew Walker}{contrib}
\icmlauthor{Judy Shen}{contrib}
\icmlauthor{Chintan Shah}{contrib}
\icmlauthor{Blake Martin}{contrib}
\icmlauthor{Aashish Sood}{contrib}
\icmlauthor{Elliot Miller}{contrib}
\icmlauthor{Ben Glass}{senior}
\icmlauthor{Andy Beck}{senior}
\icmlauthor{Harsha Pokkalla}{senior}
\icmlauthor{Syed Ashar Javed}{senior}

PathAI

% \icmlauthor{Firstname2 Lastname2}{equal,yyy,comp}
% \icmlauthor{Firstname3 Lastname3}{comp}
% \icmlauthor{}{sch}
% \icmlauthor{}{sch}
\end{icmlauthorlist}

\icmlaffiliation{lead}{Project Lead and First Author}
\icmlaffiliation{core}{Core Contributor}
\icmlaffiliation{contrib}{Contributor}
\icmlaffiliation{senior}{Senior Contributor}

\icmlcorrespondingauthor{Harshith Padigela}{harshith.padigela@pathai.com}


\vskip 0.3in
]



\printAffiliationsAndNotice{}  % leave blank if no need to mention equal contribution
% \linenumbers
% \modulolinenumbers[1]  % <-- number every line

\begin{abstract}
Foundation models trained on large-scale pathology image corpora have demonstrated strong transfer capabilities across diverse histopathology tasks. 
Building on this progress, we introduce \textbf{PLUTO-4}, our next generation of pathology foundation models that extend the Pathology-Universal Transformer (PLUTO) to frontier scale. 
We share two complementary Vision Transformer architectures in the PLUTO-4 family: a compact and efficient \textbf{PLUTO-4S} model optimized for multi-scale deployment using a FlexiViT setup with 2D-RoPE embeddings, and a frontier-scale \textbf{PLUTO-4G} model trained with a single patch size to maximize representation capacity and stability. 
Both models are pretrained using a self-supervised objective derived from DINOv2 on a large multi-institutional corpus containing 551,164 WSIs from 137,144 patients across over 50 institutions, spanning over 60 disease types and over 100 stains. 
Comprehensive evaluation across public and internal benchmarks demonstrates that PLUTO-4 achieves state-of-the-art performance on tasks requiring varying spatial and biological context, including tile classification, segmentation, and slide-level diagnosis. 
The compact PLUTO-4S provides high-throughput and robust performance for practical deployment, while PLUTO-4G establishes new performance frontiers across multiple pathology benchmarks, including 11\% improvement in dermatopathology diagnosis. These diverse improvements underscore PLUTO-4’s potential to transform real-world applications as a backbone for translational research and diagnostic use cases.
% Together, these models advance the development of scalable and general-purpose pathology foundation models designed for both research and clinical applications.
\end{abstract}

\section{Introduction}

Pathology is the study of microscopic tissue morphology and remains the clinical gold standard for diagnosing disease. 
The digitization of histopathology slides into whole slide images (WSIs) has enabled large-scale quantitative analysis 
and the development of computational pathology systems that aim to assist pathologists in diagnosis, prognosis, and biomarker assessment~\cite{campanella2019clinical, bulten2020automated}. 
However, the intrinsic complexity of WSIs --- including gigapixel scale, variations in staining protocols and scanner systems, and heterogeneous biological content --- 
poses unique challenges for machine learning methods.

Recent advances in self-supervised learning have shown that \textit{foundation models} (FMs), trained on large and diverse image corpora, 
can generate transferable visual representations across a wide range of tasks~\cite{oquab2024dinov2, caron2021emerging}. 
In pathology, several models such as PLUTO \cite{juyal2024pluto}, H-Optimus \cite{hoptimus0}, Virchow2~\cite{zimmermann2024virchow2}, 
Atlas~\cite{alber2025atlas}, 
RudolfV~\cite{dippel2024rudolfv}, 
and UNI~\cite{chen2024uni} 
have demonstrated the potential of this paradigm, 
providing encoders that generalize across tissue types, stains, and magnifications. 
Despite these advances, scaling pathology FMs remains constrained by three key factors: 
(1) limited availability of large and heterogeneous training datasets, 
(2) training instability and compute bottlenecks when training large scale models, and 
(3) high computational demands that limit deployment in diagnostic workflows at scale.

\begin{table*}[t]
\centering
\captionsetup{justification=centering}

\caption{
\textbf{Performance comparison of PLUTO-4 with existing pathology foundation models.}\\
PLUTO-4G achieves best-in-class performance for the majority of benchmarks and task categories.}
\label{tab:cls_summary}
\resizebox{\textwidth}{!}{
\begin{tabular}{lcccccccc}
\toprule
\textbf{Dataset / Metric} 
& \textbf{PLUTO-4G} 
& \textbf{H-Optimus-0} 
& \textbf{Atlas} 
& \textbf{Virchow-2} 
& \textbf{UNI2-H} 
& \textbf{Prov-Gigapath} 
& \textbf{Lunit-S} 
& \textbf{H-Optimus-1} \\
\midrule
\multicolumn{9}{l}{\textit{Spatial Transcriptomics}} \\
\multicolumn{9}{l}{\textit{(Pearson $r$)}} \\
HEST  
& \textbf{0.427} 
& 0.413 
& 0.399 
& 0.396 
& 0.414 
& 0.386 
& -- 
& 0.422 \\
\midrule
\multicolumn{9}{l}{\textit{Tile-Level Classification; EVA}} \\
\multicolumn{9}{l}{\textit{(Balanced Accuracy \%)}} \\
MHIST 
& \textbf{87.5 (0.3)} 
& 84.3 
& 85.2 
& 86.1 
& 82.4 
& 82.9 
& 78.1 
& 83.5 \\
BreakHIS 
& 81.5 (0.4) 
& 80.1 
& -- 
& 82.1 
& \textbf{85.9} 
& 82.7 
& 74.2 
& -- \\
BACH 
& \textbf{93.8 (0.5)} 
& 75.9 
& 93.1 
& 88.3 
& 91.5 
& 75.9 
& 78.3 
& -- \\
Gleason (Arvaniti)
& \textbf{79.3 (0.7)} 
& 77.0 
& -- 
& 78.3 
& 77.5 
& 72.4 
& 75.0 
& -- \\
PCAM (test) 
& \textbf{95.1 (0.1)} 
& 94.3 
& 94.9
& 93.8 
& 95.0 
& 94.5 
& 89.7 
& -- \\
CRC
& 96.4 (0.2) 
& 95.5 
& \textbf{97.0} 
& 96.7 
& 96.5 
& 95.1 
& 94.0 
& -- \\
\midrule
\multicolumn{9}{l}{\textit{Slide-level Classification}} \\
\multicolumn{9}{l}{\textit{(Balanced Accuracy \%)}} \\
PANDA-Small (test) 
& 66.8 (1.6) 
& 67.1 
& \textbf{70.0} 
& 64.6 
& 65.7 
& 65.3 
& 61.0 
& -- \\
Derm 2K* (\textit{Macro F1 \%})
& \textbf{67.1} 
& 62.8 
& -- 
& -- 
& -- 
& -- 
& -- 
& -- \\
\midrule
\multicolumn{9}{l}{\textit{Nuclear Segmentation; EVA}} \\
\multicolumn{9}{l}{\textit{DICE}} \\

MoNuSAC 
& \textbf{70.4 (0.3)} 
& 68.5
& -- 
& 66.9 
& 64.2 
& 68.0 
& 62.9 
& -- \\
CoNSep 
& \textbf{65.0 (0.1)} 
& 64.4 
& -- 
& 64.0 
& 63.0 
& 62.6 
& 60.2 
& -- \\
\bottomrule
\end{tabular}}
\\
\footnotesize 
\vspace{0.2em}
Best score per dataset in \textbf{bold}. The std across multiple runs is shown in parantheses. 
Results on EVA for external models are reported from their leaderboard or the model's paper. 
HEST results for external models are from \cite{bioptimus2025hoptimus1} or the model’s paper. 
\textit{*Derm-2K is a proprietary dataset.}
\end{table*}

To address these challenges, we developed \textbf{PLUTO-4}, a new generation of pathology foundation models designed for both scalability, efficiency and performance. 
PLUTO-4 introduces two complementary encoders:
\begin{itemize}
    \item \textbf{PLUTO-4S}, a compact, high-throughput model incorporating a \textbf{FlexiViT} backbone~\cite{beyer2023flexivit} 
    with Rotary Positional Embeddings \textbf{(RoPE)}~\cite{su2021rope} 
    for robust performance and deployment at scale.
    \item \textbf{PLUTO-4G}, a frontier-scale model trained with a single patch size (also referred as patch-token size) and positional embedding scheme, designed to maximize 
    representation capacity and model performance.
\end{itemize}

Both models were trained on a large-scale, multi-institutional corpus capturing the broad morphological and technical diversity across disease areas, staining modalities, and scanner systems, 
and which represents one of the most comprehensive collections used for pathology foundation model training.

The training methodology builds upon the original PLUTO workflow~\cite{juyal2024pluto}, 
maintaining a self-supervised pretraining framework derived from DINOv2~\cite{oquab2024dinov2} 
with multi-resolution image sampling. 
PLUTO-4 extends this framework with improved distributed training stability, higher-precision optimization, architectural improvements and extended compute scaling, enabling consistent convergence for both compact and frontier-scale architectures.

Through extensive benchmarking in tile, slide and segmentation-level tasks, 
PLUTO-4G shows state-of-the-art performance in multiple public benchmarks, 
while PLUTO-4S provides a strong and computationally efficient alternative suitable for real-world deployment. 
Together, these models advance the goal of creating general-purpose pathology foundation models capable of supporting a broad range of diagnostic and research applications.

\begin{figure*}
    \centering
 \includegraphics[width=1.0\textwidth]
    {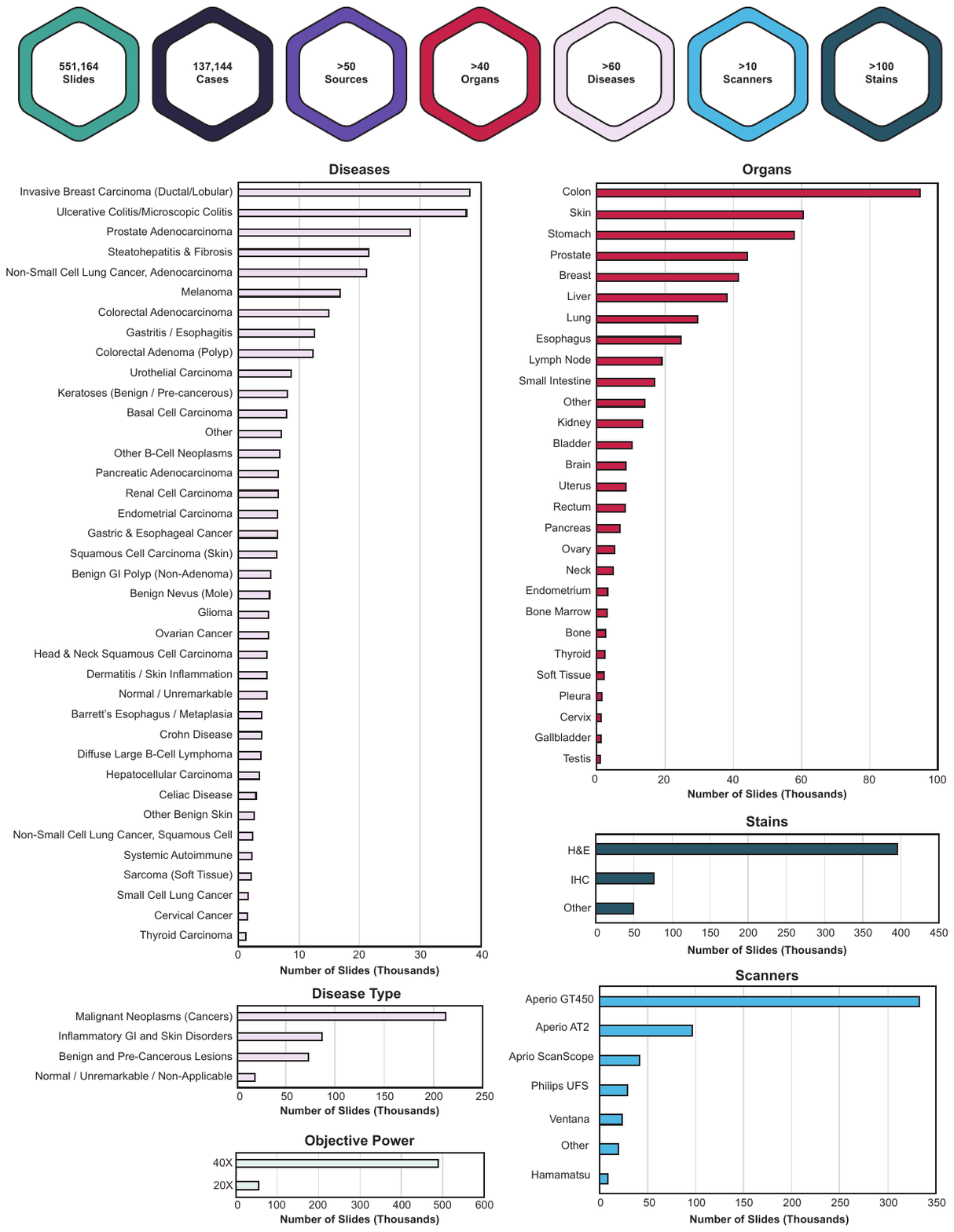}

    \caption{\textbf{Distribution of the PLUTO-4 dataset across organs, diseases, stain groups and scanners}.
    }
    \label{fig:PLUTO-4_diversity}
\end{figure*}

\section{Training Data}

\subsection{Dataset Overview}

The \textbf{PLUTO-4} dataset represents one of the largest and most diverse corpora assembled for pathology foundation model training to date. 
It comprises approximately \textbf{551,164} de-identified whole slide images(WSIs) drawn from 137,144 patients 
across more than 50 contributing institutions, spanning a broad spectrum of disease areas, organ systems, stains, and scanners. 
In total, the dataset includes slides from more than 40 distinct tissue and organ categories spanning over \textbf{60 diseases}
covering over \textbf{100 stain variants}. 
Slides were digitized on multiple scanner systems encompassing more than ten hardware models. 
The dataset reflects both diagnostic and research use cases, incorporating routine hematoxylin and eosin (H\&E) stains, 
immunohistochemistry (IHC) panels, and special stains from frozen and formalin-fixed paraffin-embedded (FFPE) preparations.

The large-scale diversity of PLUTO-4 is a critical enabler of robust and generalizable pathology foundation models. 
By integrating data across multiple healthcare systems and scanner vendors, the dataset captures a wide distribution of staining, scanning, 
and morphological variability encountered in clinical practice. 

\subsection{Stain and Modality Diversity}

As shown in Figure~\ref{fig:PLUTO-4_diversity} (bottom left), the dataset is anchored by a large cohort of approximately 396K H\&E stained WSIs, which is complemented by a diverse collection of special stains and Immunohistochemistry (IHC) slides. This IHC subset features over 50 unique targets, encompasses membranous, cytoplasmic, and nuclear staining patterns, and includes a wide range of biomarkers (e.g., PD-L1, HER2, Ki-67, CD8, CD20, AE1/AE3, ER, PR) from common to less frequent ones used in hematopathology and research applications. Special stains (e.g., Masson's trichrome, PAS, reticulin, iron) and frozen-section preparations provide additional morphological variety 
beyond routine diagnostic material. 
This stain-level diversity supports the development of encoders capable of recognizing cellular and structural features 
under heterogeneous imaging conditions.

\subsection{Scanner and Site Heterogeneity}

Slides were digitized using a variety of whole-slide scanners, including systems from Aperio, Philips, Ventana and Hamamatsu. 
Figure~\ref{fig:PLUTO-4_diversity} (bottom right) illustrates the broad vendor representation, with the largest contributions 
from Aperio and Philips scanners. 
The dataset also includes a subset of slides converted from generic TIFF formats, representing legacy or research-origin data. 
This diversity of scanner hardware and color calibration profiles introduces natural variation in image tone, contrast, 
and sharpness, which can improve feature generalization in large-scale representation learning.

\subsection{Organ and Disease Coverage}

The dataset has broad coverage across oncology and non-oncology domains, spanning over 40 organs. 
High-volume tissues such as Colon, Skin, Prostate, Breast, and Liver collectively contribute about 240K WSIs, 
with substantial representation from additional gastrointestinal, respiratory, and lymphoid organs, among others (Figure~\ref{fig:PLUTO-4_diversity}, top). 
This broad anatomic coverage ensures inclusion of both common and rare organ systems encountered in routine clinical practice.

Across these organs, the dataset encompasses more than \textbf{60 disease entities}, 
spanning malignant, benign, inflammatory, and normal tissue categories. 
Frequent malignancies include \textit{Invasive Breast Carcinoma}, \textit{Prostate Adenocarcinoma}, \textit{NSCLC Adenocarcinoma}, \textit{Colorectal Adenocarcinoma}, \textit{Melanoma}, 
and \textit{Urothelial Carcinoma}, while non-neoplastic conditions such as \textit{Ulcerative and Microscopic Colitis}, \textit{Gastritis}, and \textit{Esophagitis} 
are also represented. 
Benign and pre-cancerous lesions (\textit{Adenomas}, \textit{Keratoses}), inflammatory disorders, and unremarkable tissues further expand diagnostic diversity.

At the disease-type level (Figure~\ref{fig:PLUTO-4_diversity}, second row), 
malignant neoplasms account for approximately 212K WSIs, 
inflammatory gastrointestinal and dermatologic disorders for 86K WSIs, 
and benign or pre-cancerous conditions for 72K WSIs, 
with the remainder corresponding to normal or unclassified tissue. 
This composition provides balanced representation across major histologic categories, 
capturing variations in staining, scanning, and disease morphology observed in clinical practice.

\subsection{Data Preprocessing and Sampling}

To ensure consistent quality across magnifications, we applied a multi-stage preprocessing pipeline. 
Usable tissue regions were extracted using the latest version of our Artifact Detect algorithm \cite{Le2025ArtifactDetection}, which segments usable tissue from background and excludes artifact regions like pen markings, folds, or scanning blur.
We sampled 165M regions of usable tissue from the slides and generated around 640M image tiles across 4 magnifications (0.25, 0.5, 1.0, and 2.0~µm/px) for training with a mix of tile sizes 275px, 550px which are cropped to the global and local crops views during training.

% \subsection{Comparison to Prior Datasets}

% Compared to previous pathology foundation model datasets, PLUTO-4 represents an order-of-magnitude increase in both 
% diversity of source-sites and stain heterogeneity. 
% For reference, RudolfV~\cite{dippel2024rudolfv} utilized approximately 134,000 WSIs from two institutions, 
% Atlas~\cite{alber2025atlas} was trained on 1.2 million WSIs from Mayo Clinic and Charité, 
% and Virchow2~\cite{zimmermann2024virchow2} included 3.1 million slides from MSKCC. 
% While Virchow2 remains the largest in absolute slide count, PLUTO-4 exhibits greater cross-institutional and scanner diversity, 
% and uniquely integrates multi-stain and multi-resolution sampling from a unified corpus. 
% This combination of scale and heterogeneity is designed to improve robustness across diagnostic contexts and downstream adaptation tasks.

\section{Training Methodology}

\subsection{Training Architecture Design}

The training architecture for PLUTO-4 extends the self-supervised framework introduced in the original PLUTO~\cite{juyal2024pluto}, 
while incorporating architectural and optimization refinements that enable stable scaling from compact to frontier-sized vision transformers.

\paragraph{PLUTO-4S.}
The compact encoder PLUTO-4S is designed to support multiple levels of representational granularity across diverse pathology tasks. 
Different problems, such as cell segmentation versus slide-level classification, require varying receptive field sizes and granular context to capture relevant morphologic context. 
To accommodate this, we adopt the \textbf{FlexiViT} setup~\cite{beyer2023flexivit} similar to PLUTO \cite{juyal2024pluto}, allowing the same backbone to operate on variable patch-token sizes. 
During pretraining, patch-token sizes from [8, 16, 32] are sampled dynamically across training iterations, 
exposing the model to multi-scale input structure without separate backbone training. 
Unlike the absolute positional embeddings used in PLUTO~\cite{juyal2024pluto}, PLUTO-4S employs 
two-dimensional rotary positional embeddings \textbf{(2D-RoPE)}~\cite{su2021rope}, 
which provide relative positional encoding in both spatial dimensions and improve stability at large sequence lengths. 
This modification is consistent with the positional encoding formulation used in \cite{siméoni2025dinov3} and other large-scale vision transformers.

\paragraph{PLUTO-4G.}
For the frontier-scale model, PLUTO-4G, we observed that larger embeddings inherently possess sufficient representational capacity, and the difference between various patch-token sizes diminishes as model scale increases. 
Empirically, we saw the difference in downstream performance between patch-token sizes 8, 16 reduce across tasks as model size grows, while the computational cost and memory scales quadratically when patch-token size halves (or sequence length doubles). An example is seen in Figure \ref{fig:throughput_scaling_models_suffix_nodes}, where we see ViT-G-14 is almost 3.5X faster than ViT-G-8. Therefore, we train PLUTO-4G using a single patch-token size (14), a choice consistent with other FMs and significantly reduces our training cost while improving throughput.

\subsection{Stabilizing Self-Supervised ViT Training}

Self-Supervised training of Vision transformers with DINOv2 is unstable and challenging. The iBOT and DINO projection heads are prone to large activations and overflows, and training suffers from noisy losses, large gradient norms, and many of these become prominent as the model size grows. Similar observations were also noted by \cite{zimmermann2024virchow2}. We have found the following training choices helpful in stabilizing training in addition to the recommendations in DINOv2 like gradient clipping.

\begin{itemize}
    \item \textbf{bfloat16-mixed precision}: Even with gradient clipping we found activations in the projection heads grow and overflow in float16 which has a limited range of $[-65504, 65504]$ leading to nan losses in training. bfloat16 trades off precision for range and has a much larger range of $[-3.39 \times 10^{38}, 3.39 \times 10^{38}]$ similar to float32. While float16 may still be critical in other workloads where precision is critical like RL training \cite{qi2025defeatingtraininginferencemismatchfp16} and inference, for DINOv2 we found range more crucial than precision and thus use bfloat16 for model forward and backward passes. Key computations which need higher precision like loss computation, momentum center updates are performed in float32.

    \item \textbf{Adding register tokens}: We add in register tokens \cite{darcet2024visiontransformersneedregisters} which can capture any high norm tokens as the model size increases and better capture spatial features.
    \item \textbf{Using large batch-sizes}: We found it helpful to have batch sizes of at least 1024 to ensure stable losses and gradients. 
\end{itemize}

To ensure stable training, we use a local batch size of 32 tiles per GPU, corresponding to an effective global batch size of 1024 for PLUTO-4G. For PLUTO-4S we use an effective batch-size of 1536.
We employ the AdamW optimizer with a fixed weight decay and linear warm-up of learning rate followed by a cosine annealing decay for both models. Temperature and momentum schedules follow a similar approach to DINOv2, though we adjusted epoch lengths to ensure good coverage of the dataset.
Gradient clipping is applied at a global norm of 3.0 and we use 4 register tokens for PLUTO-4G.

\subsection{Scaling Model Training}

\begin{figure}[t]
    \centering
        \includegraphics[width=0.9\linewidth]{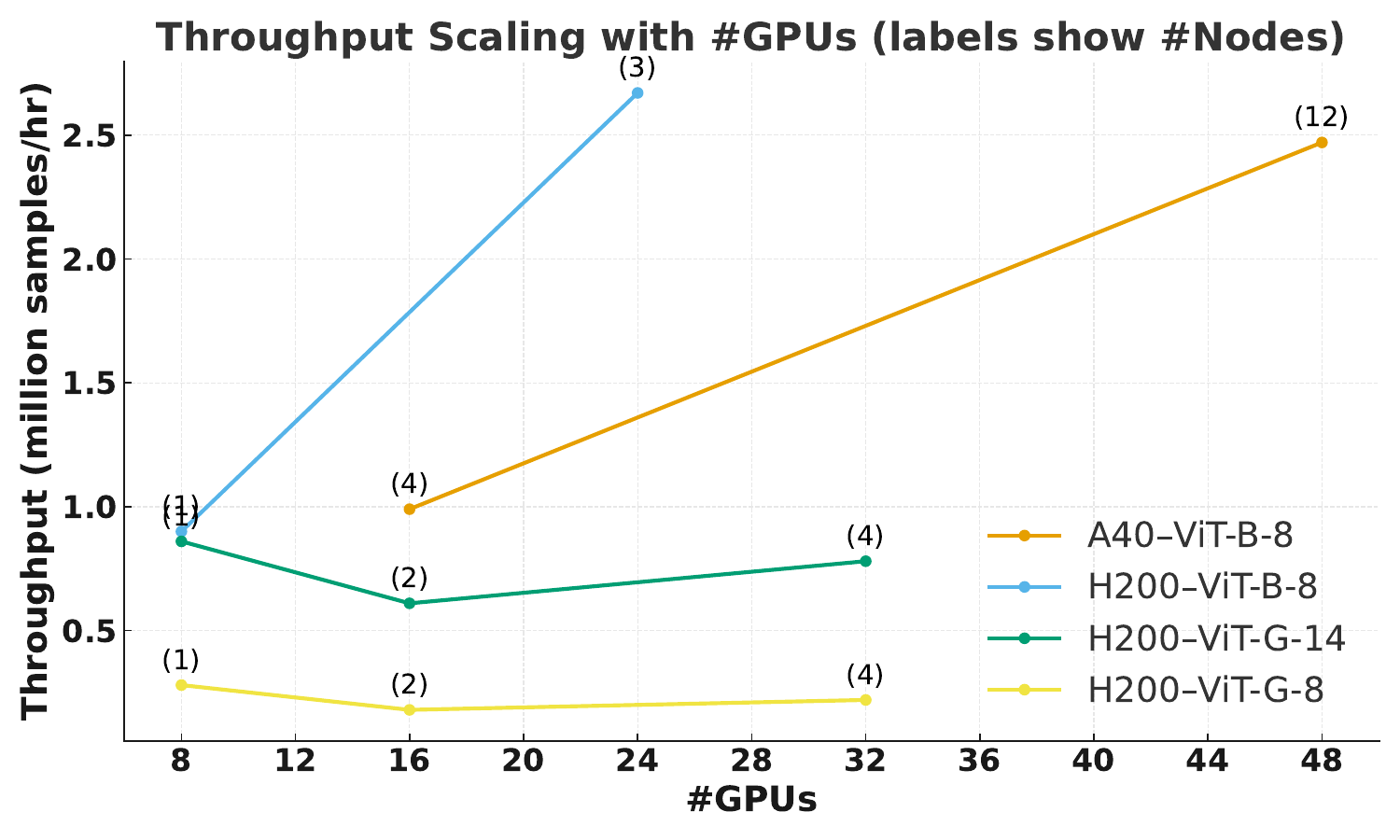}

    \caption{\textbf{Training throughput scaling across architectures and hardware.}
    \label{fig:throughput_scaling_models_suffix_nodes}
    ViT-B shows near-linear scaling across both A40 and H200 clusters, while ViT-G throughput degrades beyond two nodes due to communication bottlenecks in DDP. Additionally we can also see, ViT-G with patch-token size 8 is approximately 3.5$\times$ slower than ViT-G with patch-token size 14.}
\end{figure}

We systematically analyze scaling efficiency when training PLUTO models across multiple GPUs and nodes. Training is performed on four NVIDIA H200 nodes, each containing eight GPUs interconnected via NVLink for intra-node communication and InfiniBand with GPUDirect RDMA for inter-node communication. Figure~\ref{fig:throughput_scaling_models_suffix_nodes} summarizes the observed throughput trends for ViT-B and ViT-G architectures under varying hardware and node configurations.

\begin{figure}[t]
    \centering
    \includegraphics[width=0.9\linewidth]{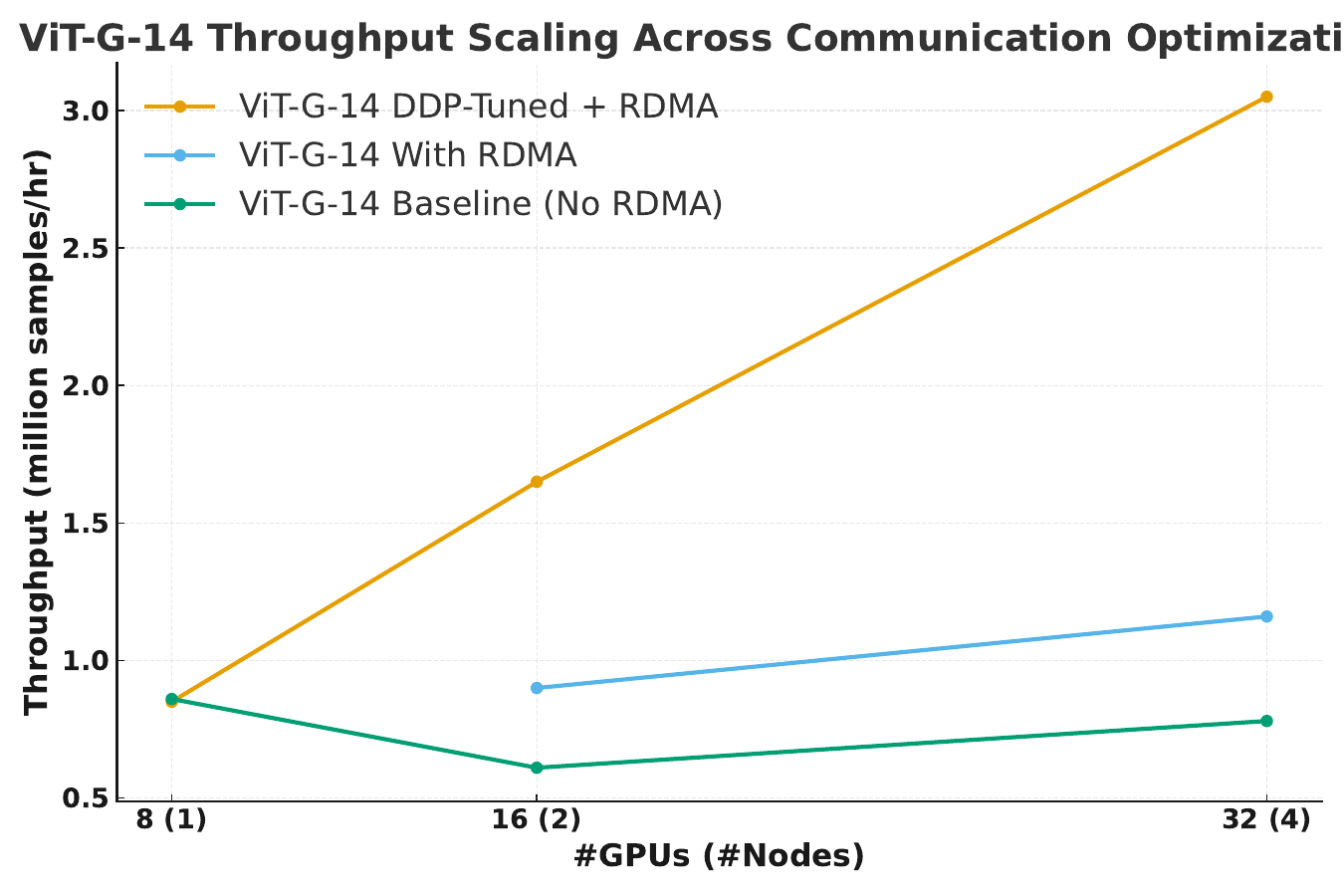}
    \caption{\textbf{Optimizing ViT-G training throughput.}
    Enabling GPUDirect RDMA and tuning DDP parameters (\texttt{bucket\_cap\_mb}, \texttt{gradient\_as\_bucket\_view}) restores near-linear throughput scaling and saturates InfiniBand bandwidth.}
    \label{fig:vitg14_rdma_ddp_scaling_nodes}
\end{figure}

For smaller architectures such as ViT-B, throughput scales almost linearly with the number of nodes on both A40 and H200 clusters. However, for larger architectures such as ViT-G, throughput drops significantly when going beyond one node, with overall throughput even decreasing at higher GPU counts. The main bottleneck arises from all-reduce synchronization during Distributed Data Parallel (DDP) training. As model size grows, the gradient tensors exchanged between GPUs increase substantially, causing network congestion and underutilization of available compute resources. This effect is particularly evident in models with large sizes (ViT-G) where per-layer gradient aggregation dominates total step time.

\paragraph{Mitigating cross-node bottlenecks.}  
To alleviate these communication bottlenecks, we enabled GPUDirect RDMA to allow direct GPU-to-GPU data transfer across nodes, bypassing CPU memory and reducing PCIe latency. While this provided an immediate improvement, scaling efficiency remained suboptimal and bandwidth utilization was still low, suggesting that the DDP communication pipeline itself required further optimization.

We conducted an extensive sweep of DDP parameters to tune gradient synchronization behavior. Two adjustments proved critical: increasing \texttt{bucket\_cap\_mb} from the default 25\,MB to 360\,MB, and enabling \texttt{gradient\_as\_bucket\_view=True}. The larger communication bucket size reduces the number of all-reduce calls per step, while \texttt{gradient\_as\_bucket\_view} minimizes tensor copies during gradient aggregation. Combined with RDMA, these modifications significantly improved communication–computation overlap, achieving near-linear throughput scaling with the number of GPUs and full utilization of the InfiniBand fabric (Fig.~\ref{fig:vitg14_rdma_ddp_scaling_nodes}).  

\paragraph{FSDP vs. DDP trade-offs.}  
We also evaluated Fully Sharded Data Parallel (FSDP) as an alternative for scaling large models. FSDP shards parameters, gradients, and optimizer states across GPUs, reducing memory and communication per device. This approach is advantageous when models exceed single-GPU VRAM capacity or when network bandwidth is constrained. However, when bandwidth is ample and models fit in GPU memory, the additional sharding and all-gather overhead reduces throughput. On H200, tuned DDP achieved roughly twice the throughput of FSDP for ViT-G. FSDP could be useful for future PLUTO variants where model scale or interconnect limits make full replication infeasible.

\section{Results and Evaluation}

\begin{table}[t]
\centering
\scriptsize
\setlength{\tabcolsep}{2pt}
\renewcommand{\arraystretch}{1.1}
\caption{\textbf{Performance results using CLS + Mean(Patch-Token) concatenated embeddings.}  
PLUTO-4G maintains state-of-the-art performance across most datasets, demonstrating strong generalization with CLS + Patch-Token embeddings.
}
\label{tab:cls+patch_summary}
\begin{tabularx}{\columnwidth}{@{}l *{4}{>{\centering\arraybackslash}X}@{}}
\toprule
\textbf{Dataset} & \textbf{PLUTO-4G} & \textbf{H-Optimus-0} & \textbf{Virchow-2} & \textbf{Atlas} \\
\midrule
HEST               & \textbf{0.432} & 0.424 & 0.398 & 0.421 \\
MHIST              & \textbf{87.9}  & 85.0  & 85.9  & 86.4  \\
BACH               & \textbf{93.2}  & 74.2  & 88.7  & 92.5  \\
PCAM (Test)        & \textbf{95.2}  & 94.3  & 93.9  & 94.8  \\
CRC         & 96.8           & 96.2  & 96.7  & \textbf{97.1} \\
PANDA-Small (Test)       & 66.6           & 68.0  & 66.4  & \textbf{70.5} \\
\bottomrule
\end{tabularx}
\footnotesize 
\vspace{0.2em}
Best result per dataset in \textbf{bold}. Results for H-Optimus-0, Virchow-2, and Atlas are from the Table~5 in \cite{alber2025atlas}. The metric for HEST is pearson r and the metric for rest of the datasets is balanced accuracy.
\end{table}

\subsection{Evaluation Methodology}

We evaluate the PLUTO-4 models on a broad set of public and proprietary benchmarks spanning multiple biological scales, 
including Tile-Level classification and regression, segmentation, and slide-level prediction. 
The evaluations were selected to assess the generality of the learned representations across tasks requiring  different contextual granularity, ranging from subcellular detail to whole-slide inference.

\paragraph{Evaluation Framework.}
We leverage the open-source \textbf{EVA} benchmarking framework \cite{kaiko.ai2024eva} for standardized evaluation of pathology foundation models. 
EVA provides unified loaders, preprocessing, and evaluation metrics across a wide collection of pathology datasets, 
allowing consistent comparison to publicly available FMs such as Virchow2~\cite{zimmermann2024virchow2}, 
UNI2-H~\cite{chen2024uni}, Prov-Gigapath ~\cite{gigapath}, Lunit~\cite{kang2023benchmarkingselfsupervisedlearningdiverse} and Atlas ~\cite{alber2025atlas}. 
For HEST we use the benchmarking setup and code as described in \cite{jaume2024hest}. Details regarding specific datasets are covered in subsequent section.

\paragraph{Embedding Selection.}
Unless otherwise specified, all reported results use the \textbf{CLS token embedding} extracted from the frozen encoder as the image representation. 
This design enables fair comparison to prior works that primarily rely on CLS embeddings for linear probing and downstream task evaluation. 
Results using the concatenation of CLS, Mean of patch-token embeddings for classification tasks are provided in the table \ref{tab:cls+patch_summary}
for completeness and to quantify the effect of incorporating spatial context.

\subsection{Evaluation Datasets and Reporting Protocols}

We evaluate PLUTO-4 models across four broad task categories representing different biological and spatial scales in pathology: Tile-Level classification, segmentation, spatial transcriptomics, and slide-level prediction. For all benchmarks, we follow standard linear-probing or ABMIL protocols as implemented in EVA~\cite{kaiko.ai2024eva} and HEST~\cite{jaume2024hest}.

\paragraph{Tile-Level Classification.}
We evaluate on MHIST, BreakHIS, BACH, and Gleason benchmarks from the EVA suite, which span epithelial tissue classification and breast cancer subtyping tasks requiring moderate contextual resolution. 
Metrics are reported as Balanced Accuracy (BA) averaged over 5 runs using the EVA evaluation protocol.
For comparison with external models, results for H-Optimus-0, Virchow-2, UNI2-H, Prov-Gigapath, and Lunit are taken directly from the \texttt{EVA leaderboard}, while Atlas and H-Optimus-1 results are taken from their respective publications~\cite{alber2025atlas,bioptimus2025hoptimus1}.

\paragraph{Segmentation.}
We benchmark on CoNSep and MoNuSAC datasets for instance-level nuclei segmentation and classification across organs and stains. We use the $ConvDecoderWithImage$ decoder from EVA which upsamples the input feature maps before concatenating them with the original input image, followed by a series of convolutional layers.  
Performance is reported is the MONAI-DICE score averaged over 5 EVA evaluation runs.
Results for H-Optimus-0, Virchow-2, UNI2-H, Prov-Gigapath, and Lunit are taken from the EVA leaderboard. 
Segmentation metrics for Atlas and H-Optimus-1 are not reported in their respective works.

\paragraph{Spatial Transcriptomics (HEST).}
We use the HEST-1k benchmark~\cite{jaume2024hest} to evaluate spatially-resolved gene expression prediction from H\&E morphology across nine tissue types.
We follow the official Ridge Regression with PCA protocol and report the average Pearson correlation across all tissue types.
Results for all external foundation models are drawn from the unified comparison presented in the H-Optimus-1 blog ~\cite{bioptimus2025hoptimus1}, and results for Atlas are cited from its publication~\cite{alber2025atlas}. 
Indication-wise results are provided in Table~\ref{tab:hest_results}.

\paragraph{Slide-level Prediction.}
Two slide-level tasks are used to assess global contextual reasoning:
\begin{enumerate}[label=(\roman*)]
    \item \textbf{PANDA-Small:} A subset of the PANDA challenge,  for predicting prostate cancer Gleason grades, using the EVA ABMIL protocol. We report \textbf{Balanced Accuracy} averaged over 20 runs using EVA.
    \item \textbf{Derm-2K:} A proprietary dermatopathology diagnosis benchmark comprising over 2,000 whole-slide images across 17 skin lesion categories. The dataset covers major lesion types, including actinic keratosis, basal cell carcinoma, benign nevus, cyst, dermatitis, dysplastic nevus, invasive melanoma, lichenoid keratosis, melanoma in situ, scar, seborrheic keratosis, squamous cell carcinoma, squamous cell carcinoma in situ, vascular lesion, verruca vulgaris, other benign non-melanocytic lesions, and normal skin. We evaluate models using our internal AB-MIL setup and report \textbf{Macro-F1} score.
\end{enumerate}
For PANDA-Small, results for H-Optimus-0, Virchow-2, UNI2-H, Prov-Gigapath, and Lunit are taken from the EVA leaderboard; Atlas results are reported from~\cite{alber2025atlas}. 
For Derm-2K, results are compared between PLUTO-4G and H-Optimus-0.

\subsection{Benchmark Results}
\subsubsection{Tile-Level Classification}

PLUTO-4G demonstrates strong generalization on a diverse set of Tile-Level classification tasks 
(Table~\ref{tab:cls_summary}). 
On MHIST, PCAM-test and BACH, \textbf{PLUTO-4G} achieves 
the highest balanced accuracy outperforming other comparable models like Virchow-2, Atlas, H-Optimus-0 and H-Optimus-1. 
For fine-grained glandular structure classification (Gleason Arvaniti), PLUTO-4G attains the highest accuracy, indicating its ability to model large-scale context. 
Across general-purpose benchmarks such as BreakHIS and CRC, PLUTO-4G achieves top-tier accuracy, establishing a new performance standard for frozen representation evaluation.

\subsubsection{Segmentation}
On dense nuclear segmentation and classification tasks (MoNuSAC and CoNSep), PLUTO-4G attains the highest Dice and AJI scores 
(70.4 and 65.0, respectively), outperforming all previous pathology foundation models, 
including Virchow-2 (0.669 / 0.640) and UNI2-H (64.2 / 63.0). 
These gains demonstrate that self-supervised representations from PLUTO-4 maintain spatial coherence, enabling effective adaptation for dense instance segmentation.

\subsubsection{Spatial Transcriptomics (HEST)}

The HEST benchmark evaluates morphological correlates of spatially resolved gene expression across nine tumor types. 
As shown in Table~\ref{tab:hest_results}, PLUTO-4G achieves the highest mean Pearson correlation 
($r = 0.427$), surpassing all prior models including H-Optimus-1 (0.422), UNI2-H (0.413), Virchow-2 (0.396), 
and Atlas (0.399). 
PLUTO-4G shows particularly strong improvements in the clear cell renal cell carcinoma (CCRCC), pancreatic ductal adenocarcinoma (PAAD), and skin cutaneous melanoma (SKCM) datasets, 
indicating enhanced sensitivity to morpho-molecular associations across distinct tumor morphologies.

\subsubsection{Slide-Level Prediction}

We evaluate PLUTO-4G on two slide-level tasks requiring aggregation of regional features into whole-slide predictions.

On the PANDA-Small Gleason grading benchmark, PLUTO-4G demonstrates strong performance, surpassing models such as Virchow-2, UNI2-H, and Prov-Gigapath, performing on par with H-Optimus-0, and slightly below Atlas.(Table~\ref{tab:cls_summary}). 

For Derm-2K, we evaluate PLUTO-4G and H-Optimus-0 on a proprietary 17-class dataset of 2K WSIs
covering benign, inflammatory, and malignant skin diseases. 
PLUTO-4G achieves a macro-F1 of 67.1, surpassing H-Optimus-0 (0.628). 

These results demonstrate robust transfer of learned morphological features to slide-level prediction prediction tasks.

\newcolumntype{C}{>{\centering\arraybackslash}X}

\begin{table*}[t]
\centering
\small % <- lock the font size (or \normalsize/\footnotesize)
\caption{\textbf{Performance results for PLUTO-4: CLS vs.\ CLS + Mean(Patch-Token) across patch-token sizes and models.}
ViT-S models trained with FlexiViT achieve strong performance while allowing configuration of the best patch-token size [8 / 16] for the task. Patch-Token size 8 significantly improves 4S performance on tasks needing fine-grained context, while patch-token size of 16 can perform on-par or better than patch-token size 8 on many coarse grained tasks while being 4X faster. \\ Using CLS + Patch embeddings yields modest improvements across most classification tasks.  
}

\label{tab:cls_vs_patch_ps}
% set table width explicitly (keeps font constant and centers the block)
\begin{tabularx}{0.95\textwidth}{@{}lCCC CCC@{}}
\toprule
 & \multicolumn{3}{c}{\textbf{CLS token only}} & \multicolumn{3}{c}{\textbf{CLS + Mean(Patch-Token)}} \\
\cmidrule(lr){2-4}\cmidrule(lr){5-7}
\textbf{Dataset / Metric} & \textbf{4S-8} & \textbf{4S-16} & \textbf{4G} & \textbf{4S-8} & \textbf{4S-16} & \textbf{4G} \\
\midrule
\multicolumn{7}{l}{\textit{Spatial Transcriptomics}} \\
\multicolumn{7}{l}{\textit{(Pearson $r$)}} \\
HEST         & 0.365 & 0.362 & 0.427 & 0.369 & 0.364 & \textbf{0.432} \\
\midrule
\multicolumn{7}{l}{\textit{Tile-Level Classification}} \\
\multicolumn{7}{l}{\textit{(Balanced Accuracy \%)}} \\
MHIST                      & 83.7 & 83.5 & 87.5 & 84.2 & 83.4 & \textbf{87.9} \\
BreakHIS                   & 81.3 & 79.2 & 81.5 & 80.8 & 76.8 & \textbf{81.8} \\
BACH                       & 82.7 & 79.6 & \textbf{93.8} & 85.1 & 79.8 & 93.2 \\
Gleason (Arvaniti)         & 76.2 & 76.3 & \textbf{79.3} & 76.3 & 76.6 & 78.5 \\
PCAM (Test)                & 90.7 & 90.9 & 95.1 & 91.4 & 91.6 & \textbf{95.2} \\
CRC      & 95.2 & 95.0 & 96.4 & 95.4 & 95.5 & \textbf{96.8} \\
\midrule
\multicolumn{7}{l}{\textit{Slide-level Classification}} \\
\multicolumn{7}{l}{\textit{(Balanced Accuracy \%)}} \\
PANDA-Small (Test)               & 61.8 & 61.5 & \textbf{66.8} & 63.1 & 63.0 & 66.6 \\
% Derm-2K (Macro-F1 \%)         & --   & 62.8 & \textbf{67.1} & -- & -- & -- \\
\midrule
\multicolumn{7}{l}{\textit{Nuclear Segmentation (DICE)}} \\
MoNuSAC              & 67.8 & 64.0 & \textbf{70.4} & -- & -- & -- \\
CoNSep               & 64.9 & 62.1 & \textbf{65.0} & -- & -- & -- \\
\bottomrule
\end{tabularx}

\footnotesize 
\vspace{0.2em}
Benchmark datasets are setup and evaluated using EVA, with the exception of HEST.
For segmentation datasets, CLS + Mean(Patch-Token) features are not applicable, as segmentation decoders utilize the entire set of Patch-Token features rather than pooled representations.\end{table*}

\subsubsection{CLS + Mean(Patch-Token) Concatenation}
\label{sec:cls_patch_concat}

To ensure completeness and comparability with prior works, we additionally report results using 
CLS + Mean(Patch-Token) embeddings (Table~\ref{tab:cls+patch_summary}). 
This evaluation mirrors the setup used in Virchow2 and Atlas. Since not all prior FM papers reported these evaluations, we use results reported in Atlas for our comparison.

Across eight benchmark datasets, PLUTO-4G achieves or matches state-of-the-art performance on nearly every task, 
outperforming or equaling other external models such as Atlas, Virchow-2, and H-Optimus-0. 
Specifically, PLUTO-4G attains the highest correlation on HEST (0.432) and top accuracy on 
MHIST (87.9), BACH (93.2), and PCAM (95.2). 
For the slide-level Gleason grading PANDA-Small, PLUTO-4G achieves comparable performance to other models, and slightly below Atlas. 
Overall, these results demonstrate that PLUTO-4G’s representations remain highly competitive even when evaluated 
under alternative embedding strategies used by other pathology foundation models.

\subsection{Comparison of PLUTO Models}
\label{sec:pluto_model_comparison}

We next analyze the impact of \textbf{patch-token size} (also referred as patch size) and \textbf{embedding type} across PLUTO variants 
(Table~\ref{tab:cls_vs_patch_ps}). 
This comparison highlights differences between the smaller PLUTO-4S configurations 
(trained with multiple patch-token sizes using FlexiViT) and the larger PLUTO-4G model (trained with a single patch-token size of 14).

\paragraph{Effect of patch size (ps = 8 vs ps = 16).}
Within the ViT-S family, performance differences across patch-token sizes are modest but consistent. 
Models trained with smaller patch-tokens (ps = 8) slightly outperform ps = 16 on fine-grained, texture-rich datasets such as MHIST, BreakHIS, and BACH, with both CLS, CLS+Mean(Patch-Token) settings, reflecting improved sensitivity to local morphological cues. 
Conversely, ps = 16 performs on par or better for larger-context tasks such as Gleason, CRC and PCAM. The gap however increases in nuclear segmentation tasks CoNSep and MoNuSAC, where smaller patch-sizes can better capture granular local features with PLUTO-4S-8 performing comparably or better than many larger external FMs like Virchow-2, UNI-2H and closing the gap with PLUTO-4G.  
These results confirm that flexible patch-token training effectively balances global and local context while providing flexibility to select the best patch-token size for modelling depending on the task.

\paragraph{Scaling to ViT-G.}
The larger PLUTO-4G model delivers substantial gains across all task categories. 
It surpasses both ViT-S configurations on all Tile-Level benchmarks and achieves 
the best results on segmentation, with Dice/AJI scores of 70.4/65.0, 
indicating stronger spatial consistency in feature maps. 
This scaling trend mirrors the improvements observed in other large-scale vision transformers, 
where increased capacity enhances representational generality across tasks.

\paragraph{CLS + Mean(Patch-Token) Concatenation.}
To assess whether incorporating broader spatial context improves classification performance, 
we evaluate concatenated embeddings combining the CLS token with the mean of all patch-tokens 
(Table~\ref{tab:cls_vs_patch_ps}). 
This formulation allows the model to jointly leverage global and local contextual features during downstream evaluation. 

For PLUTO-4G, we observe moderate gains on several classification tasks, including 
HEST (+0.005 $r$), MHIST (+0.4 pp), PCAM, and CRC (+0.4 pp), 
while datasets such as BACH and Gleason show marginal declines. 
Across model variants, the improvements are not uniform but indicate a consistent trend: 
performance increases on 7 of 8 datasets for PLUTO-4S-8, 5 of 8 for PLUTO-4S-16, and 5 of 8 for PLUTO-4G. 
These results suggest that incorporating mean patch-token information can enhance discriminative ability 
for certain tasks, particularly those requiring finer-grained morphological context. 

For segmentation, CLS+Mean Patch-Token features is not directly applicable, as downstream decoders directly operate on the entire Patch-Token features. 

Overall, the addition of Tile-Level context yields modest but measurable benefits for classification tasks, particularly in smaller ViT-S models, while larger ViT-G 
models appear less sensitive to the choice of embedding aggregation strategy.

\begin{table}[t]
\centering
\setlength{\tabcolsep}{6pt}
\renewcommand{\arraystretch}{1.1}
\caption{\textbf{Derm-2K Slide-Level Prediction results.} 
PLUTO-4G achieves a 11\% relative improvement in macro-F1 over the previous PLUTO-3 series, with PLUTO-4S yielding strong improvements}
\label{tab:derm_results}
\begin{tabular}{lcc}
\toprule
\textbf{Model} & \textbf{Architecture / Size} & \textbf{Macro F1} \\
\midrule
PLUTO-3S-16 & ViT-S (22M) & 0.606 \\
PLUTO-4S-16 & ViT-S (22M) & 0.628 \\
PLUTO-4G & ViT-G (1.1B) & \textbf{0.671} \\
H-Optimus-0 & ViT-G (1.1B) & 0.628 \\
\bottomrule
\end{tabular}
\end{table}

\begin{table*}[t]
\centering
\caption{\textbf{HEST Benchmark: Spatial Transcriptomics Prediction Across Tumor Types.}
} 
\label{tab:hest_results}
% \begin{tabular}{lcccccccccc}
% \toprule
% \textbf{Model} & \textbf{CCRCC} & \textbf{COAD} & \textbf{IDC} & \textbf{LUNG} & \textbf{LYMPH} & \textbf{PAAD} & \textbf{PRAD} & \textbf{READ} & \textbf{SKCM} & \textbf{AVG} \\
% \midrule
% \textbf{PLUTO-4G} & \textbf{0.289 (0.042)} & 0.316 (0.016) & \textbf{0.606 (0.085)} & 0.569 (0.028) & 0.273 (0.048) & \textbf{0.511 (0.049)} & 0.374 (0.034) & 0.233 (0.033) & \textbf{0.670 (0.045)} & \textbf{0.427} \\
% H-Optimus-1 & 0.245 (0.125) & \textbf{0.320 (0.016)} & 0.602 (0.081) & \textbf{0.578 (0.012)} & \textbf{0.277 (0.039)} & 0.496 (0.051) & \textbf{0.378 (0.012)} & \textbf{0.242 (0.015)} & 0.659 (0.048) & 0.422  \\
% H-Optimus-0 & 0.255 (0.135) & 0.309 (0.000) & 0.598 (0.085) & 0.559 (0.032) & 0.259 (0.040) & 0.491 (0.040) & 0.385 (0.000) & 0.222 (0.048) & 0.645 (0.062) & 0.413  \\
% UNI2-H & 0.261 (0.132) & 0.301 (0.004) & 0.590 (0.081) & 0.558 (0.014) & 0.272 (0.040) & 0.500 (0.040) & 0.357 (0.049) & 0.223 (0.038) & 0.659 (0.017) & 0.413  \\
% Atlas & 0.278 (0.036) & 0.259 (0.031) & 0.596 (0.081) & 0.570 (0.017) & 0.257 (0.047) & 0.507 (0.072) & 0.353 (0.032) & 0.213 (0.029) & 0.562 (0.050) & 0.399 \\
% Virchow-2 & 0.257 (0.123) & 0.259 (0.016) & 0.592 (0.08) & 0.553 (0.017) & 0.255 (0.026) & 0.472 (0.065) & 0.348 (0.031) & 0.209 (0.050) & 0.619 (0.028) & 0.396  \\
% \bottomrule
% \end{tabular}
% }
\begin{tabular}{lcccccc}
\toprule
\textbf{Dataset} & \textbf{PLUTO-4G} & \textbf{H-Optimus-1} & \textbf{H-Optimus-0} & \textbf{UNI2-H} & \textbf{Atlas} & \textbf{Virchow-2} \\
\midrule
CCRCC & \textbf{0.289 (0.042)} & 0.245 (0.125) & 0.255 (0.135) & 0.261 (0.132) & 0.278 (0.036) & 0.257 (0.123) \\
COAD & 0.316 (0.016) & \textbf{0.320 (0.016)} & 0.309 (0.000) & 0.301 (0.004) & 0.259 (0.031) & 0.259 (0.016) \\
IDC & \textbf{0.606 (0.085)} & 0.602 (0.081) & 0.598 (0.085) & 0.590 (0.081) & 0.596 (0.081) & 0.592 (0.080) \\
LUNG & 0.569 (0.028) & \textbf{0.578 (0.012)} & 0.559 (0.032) & 0.558 (0.014) & 0.570 (0.017) & 0.553 (0.017) \\
LYMPH & 0.273 (0.048) & \textbf{0.277 (0.039)} & 0.259 (0.040) & 0.272 (0.040) & 0.257 (0.047) & 0.255 (0.026) \\
PAAD & \textbf{0.511 (0.049)} & 0.496 (0.051) & 0.491 (0.040) & 0.500 (0.040) & 0.507 (0.072) & 0.472 (0.065) \\
PRAD & 0.374 (0.034) & \textbf{0.378 (0.012)} & 0.385 (0.000) & 0.357 (0.049) & 0.353 (0.032) & 0.348 (0.031) \\
READ & 0.233 (0.033) & \textbf{0.242 (0.015)} & 0.222 (0.048) & 0.223 (0.038) & 0.213 (0.029) & 0.209 (0.050) \\
SKCM & \textbf{0.670 (0.045)} & 0.659 (0.048) & 0.645 (0.062) & 0.659 (0.017) & 0.562 (0.050) & 0.619 (0.028) \\
\midrule
AVG & \textbf{0.427} & 0.422 & 0.413 & 0.413 & 0.399 & 0.396 \\
\bottomrule
\end{tabular}

\footnotesize
\vspace{0.2em}
Performance reported as mean Pearson correlation ($r$) across folds with standard deviation in parentheses.
Bold values denote best performance per tumor type. Numbers for Atlas are extracted from their publication \cite{alber2025atlas} and other external models are extracted from \cite{bioptimus2025hoptimus1}

\end{table*}

% Requires: \usepackage{booktabs,tabularx,makecell}
% Requires: \usepackage{booktabs,tabularx,makecell}
\subsection{Case Study: Impact on PathAssist Derm Product}

% Requires: \usepackage{booktabs,makecell}

PLUTO-4 represents a significant leap in performance from our prior versions and we see these improvements across a range of our product offerings. As an example, we present results on our proprietary dermatopathology diagnosis dataset, comprising 17 lesion classes. The prior version of PLUTO corresponds to the model deployed in our PathAssist Derm product \cite{pathai_pathassist_derm}. As shown in Table \ref{tab:derm_results}, the new PLUTO-4 models achieve marked improvements, with PLUTO-4G delivering an 11\% performance boost and PLUTO-4S achieving competitive results comparable to H-Optimus-0, despite being substantially smaller in scale.

% \subsection{Discussion}

\section{Discussion}

The PLUTO-4 series demonstrates the next step in scaling and adapting foundation models for digital pathology. 
Our results establish \textbf{PLUTO-4G} as a new performance standard across diverse pathology benchmarks, 
achieving state-of-the-art results in Tile-Level, segmentation, and slide-level prediction tasks. 
Trained at scale with a large, diverse corpus of histopathology data, PLUTO-4G exhibits strong generalization across 
datasets and clinical contexts, capturing both cellular- and tissue-level morphology within a unified representation. 
In parallel, \textbf{PLUTO-4S} provides an efficient and versatile alternative—offering competitive performance 
with substantially reduced compute requirements and configurable patch-token sizes that make it well suited for 
a wide range of downstream problems and deployment settings.

Foundation models like PLUTO-4 represent a powerful substrate for computational pathology, providing general-purpose visual representations which are useful for a wide-variety of downstream tasks.
However, they are only one part of an end-to-end computational pathology system. Building useful real-world applications requires additional task-specific adaptation layers that translate general embeddings into downstream applications. Developing robust and interpretable adapters remains a key step toward making these models practically deployable.

There also exists an inherent trade-off between deployability and performance. 
Larger architectures such as PLUTO-4G deliver the higher accuracy and generalization but require greater computational resources. 
Smaller models such as PLUTO-4S offer faster inference, lower memory footprint, and easier fine-tuning while maintaining competitive accuracy. 
Together, the two variants define a scalable family of models that balance performance and accessibility, 
supporting both high-throughput research applications and real-world deployment.

Beyond research benchmarks, foundation models such as PLUTO-4 can directly elevate upstream performance across PathAI’s diagnostic and biopharma product lines. \footnote{PathExplore, IHCExplore, TumorDetect, AIM-PD-L1, AIM-HER2, AIM-TumorCellularity, and PathAssist Derm are For Research Use Only. Not for use in diagnostic procedures.}
When integrated into products such as Explore product line for biomarker discovery (e.g., PathExplore, IHCExplore), Detect and Assist product line for workflow efficiencies (e.g., TumorDetect, PathAssist Derm), and AIM products for automated, reproducible biomarker quantification (e.g., AIM-PD-L1, AIM-HER2, AIM-TumorCellularity),
these models provide stronger feature representations and significantly higher starting points for task-specific supervised fine-tuning. PathAssist Derm for reference, saw a 11\% improvement in accuracy and robustness when upgrading to PLUTO-4G. 
By serving as the representational backbone across these product lines, 
PLUTO-4 enables the development of more powerful diagnostic and research applications that accelerate biomarker discovery, 
enhance translational insights, and expand the scope of computational pathology-driven discovery.

\section*{Acknowledgements}
The authors would like to thank Ylaine Gerardin and Jacqueline Brosnan-Cashman for their valuable feedback on the manuscript. 
The authors also acknowledge Chris Burke, Geetika Singh, and Darren Fahy for their support with data onboarding and management; 
Jon Ross and Braden Senst for their MLOps and infrastructure support; and Benjamin Chen for his pathology expertise and guidance.

% \section*{Impact Statement}

% Authors are \textbf{required} to include a statement of the potential 
% broader impact of their work, including its ethical aspects and future 
% societal consequences. This statement should be in an unnumbered 
% section at the end of the paper (co-located with Acknowledgements -- 
% the two may appear in either order, but both must be before References), 
% and does not count toward the paper page limit. In many cases, where 
% the ethical impacts and expected societal implications are those that 
% are well established when advancing the field of Machine Learning, 
% substantial discussion is not required, and a simple statement such 
% as the following will suffice:

% ``This paper presents work whose goal is to advance the field of 
% Machine Learning. There are many potential societal consequences 
% of our work, none which we feel must be specifically highlighted here.''

% The above statement can be used verbatim in such cases, but we 
% encourage authors to think about whether there is content which does 
% warrant further discussion, as this statement will be apparent if the 
% paper is later flagged for ethics review.

% % In the unusual situation where you want a paper to appear in the
% % references without citing it in the main text, use \nocite
\clearpage
\bibliography{main_paper}
\bibliographystyle{icml2025}

% %%%%%%%%%%%%%%%%%%%%%%%%%%%%%%%%%%%%%%%%%%%%%%%%%%%%%%%%%%%%%%%%%%%%%%%%%%%%%%%
% %%%%%%%%%%%%%%%%%%%%%%%%%%%%%%%%%%%%%%%%%%%%%%%%%%%%%%%%%%%%%%%%%%%%%%%%%%%%%%%
% % APPENDIX
% %%%%%%%%%%%%%%%%%%%%%%%%%%%%%%%%%%%%%%%%%%%%%%%%%%%%%%%%%%%%%%%%%%%%%%%%%%%%%%%
% %%%%%%%%%%%%%%%%%%%%%%%%%%%%%%%%%%%%%%%%%%%%%%%%%%%%%%%%%%%%%%%%%%%%%%%%%%%%%%%
% \newpage
% \appendix
% \onecolumn

% You can have as much text here as you want. The main body must be at most $8$ pages long.
% For the final version, one more page can be added.
% If you want, you can use an appendix like this one.  

% The $\mathtt{\backslash onecolumn}$ command above can be kept in place if you prefer a one-column appendix, or can be removed if you prefer a two-column appendix.  Apart from this possible change, the style (font size, spacing, margins, page numbering, etc.) should be kept the same as the main body.
%%%%%%%%%%%%%%%%%%%%%%%%%%%%%%%%%%%%%%%%%%%%%%%%%%%%%%%%%%%%%%%%%%%%%%%%%%%%%%%
%%%%%%%%%%%%%%%%%%%%%%%%%%%%%%%%%%%%%%%%%%%%%%%%%%%%%%%%%%%%%%%%%%%%%%%%%%%%%%%

\end{document}